\documentclass{article}

\usepackage{PRIMEarxiv}

\usepackage[utf8]{inputenc} % allow utf-8 input
\usepackage[T1]{fontenc}    % use 8-bit T1 fonts
\usepackage{hyperref}       % hyperlinks
\usepackage{url}            % simple URL typesetting
\usepackage{booktabs}       % professional-quality tables
\usepackage{amsfonts}       % blackboard math symbols
\usepackage{nicefrac}       % compact symbols for 1/2, etc.
\usepackage{microtype}      % microtypography
\usepackage{lipsum}
\usepackage{fancyhdr}       % header
\usepackage{graphicx}       % graphics
\usepackage{graphicx}%
\usepackage{multirow}%
\usepackage{amsmath,amssymb,amsfonts}%
\usepackage{amsthm}%
\usepackage{mathrsfs}%
\usepackage[title]{appendix}%
\usepackage{xcolor}%
\usepackage{textcomp}%
\usepackage{manyfoot}%
\usepackage{booktabs}%
\usepackage{algorithm}%
\usepackage{algorithmicx}%
\usepackage{algpseudocode}%
\usepackage{listings}%
\usepackage{graphicx}
\usepackage{caption}
\usepackage{subcaption}
\usepackage{float} % Add this to the preamble for table placement control
\usepackage{subcaption}
\usepackage{bbm}

\usepackage{graphicx}
\usepackage{url}
\usepackage{amsmath,amssymb,mathtools,bm,etoolbox}
\usepackage{IEEEtrantools}
\usepackage{tabularx,varwidth,multirow,booktabs,makecell}
\usepackage{longtable,adjustbox}
\usepackage{relsize}
\usepackage{natbib}
\hypersetup{
    colorlinks=true, % Enable colored links
    linkcolor=blue,  % Color for internal links (e.g., table of contents)
    citecolor=blue,  % Color for citations
    urlcolor=blue    % Color for URLs
}
\usepackage{siunitx}
\usepackage{amsthm}
\theoremstyle{definition} 
\usepackage{xcolor}%

\usepackage{subcaption}
\usepackage{listings}

\usepackage{listings}
\usepackage{fancybox} % For the framed box

\theoremstyle{thmstyleone}%
\theoremstyle{thmstyletwo}%

\theoremstyle{thmstylethree}%

\title{Automated explanation of machine learning models of footballing actions in words
}

\author{
  Pegah Rahimian\\
  Dept. of Information Technology \\
  Uppsala University \\
  \texttt{pegah.rahimian@it.uu.se} \\
   \And
  Jernej Flisar \\
  Twelve Football \\
  \texttt{jernej@twelve.football} \\
   \And
  David Sumpter \\
  Dept. of Information Technology \\
  Uppsala University \\
  Twelve Football \\
  \texttt{david.sumpter@it.uu.se} \\ \\
}

\begin{document}
\maketitle

\begin{abstract}
While football analytics has changed the way teams and analysts assess performance, there remains a communication gap between machine learning practice and how coaching staff talk about football. Coaches and practitioners require actionable insights, which are not always provided by models. To bridge this gap, we show how to build wordalizations (a novel approach that leverages large language models) for shots in football. Specifically, we first build an expected goals model using logistic regression. We then use the co-efficients of this regression model to write sentences describing how factors (such as distance, angle and defensive pressure) contribute to the model's prediction. Finally, we use large language models to give an entertaining description of the shot. We describe our approach in a model card and provide an interactive open-source application describing  shots in recent tournaments. We discuss how shot wordalisations might aid communication in coaching and football commentary, and give a further example of how the same approach can be applied to other actions in football.
\end{abstract}

% keywords can be removed
\keywords{Soccer Analytics \and Explainable AI \and Expected Goal \and Language Models}

\section{Introduction}\label{sec1}
The field of soccer analytics has witnessed a rapid evolution, with machine learning models playing a crucial role in evaluating player and team performance \cite{Tom2019, gyarmati2016qpass, Fernandez2020, Fernandez2019, pegah2022, pegah2023-2, Brefeld2022}. One of the most widely used models is the Expected Goals (xG) model, which is used to evaluate the quality of scoring opportunities by assigning probabilities to shots based on factors such as location, angle, and defensive pressure \cite{pollard1997measuring,sumpter2016soccermatics}. Several studies train machine learning models using predictors such as shot type, distance to goal, and angle to goal to estimate xG (e.g., \cite{Rathke2017AnEO, Tippana2020, Pardo2020, Herbinet2018, Wheatcroft2021, Bransen2021, Sarkar2021, Eggels2016}). Additionally, detailed studies have investigated the role of data sources on model performance \cite{Davis2020}; how defensive positioning and goalkeeper placement enhances the estimation of goal probabilities \cite{lucey2015quality}; and the use of neural networks to estimate scoring probabilities \cite{Ruiz2015MeasuringSE}.

An important consideration when building xG models is that we should be able to explain their implications to coaching staff. Many xG models are black boxes, producing numerical probabilities without offering clear explanations of how the different features of a shot determine the probability that it will result in a goal \cite{davis2024challenges}. To address this challenge, one approach is to use SHAP (SHapley Additive Explanations) \cite{lundberg2017unified} to explain the contribution of each feature to a model's prediction. \cite{Anzer2021AGS} applied SHAP to xG models, demonstrating that shot distance is the most influential factor in goal probability. Another approach is to build models that are interpretable by design. For example, building on work by Morales and Sumpter \cite{morales2016mathematics}, \cite{sumpter2016soccermatics} proposes a logistic regression model that incorporates how much of the goal the shooter can see and distance from goal as variables. This expected goals model can thus be explained in terms of the shooter sight on goal, a simple to communicate coaching concept. 

Even when adopting these approaches, their remains a gap between what a machine learner practitioner and coaching staff might consider as an explanation. Indeed, while SHAP values give a numerical representation of feature contributions, these do not automatically translate into actionable insights for football practitioners. This is part of a larger issue within sports analytics where very few studies explain how adopting recommendations from a model impact performance \cite{goes2021unlocking}. To bridge this gap, we adapt an approach introduced by \cite{wordalisation} known as \textit{Wordalisation}. The key idea is is to use large language models (LLMs) to convert numbers into natural language narratives. One example in \cite{wordalisation} is a football scout, which uses rankings of players in key metrics to describe their skills. Wordalisations are thus concise, easily digestible narratives that summarize data-driven observations without directly reporting numerical values. Prompt engineering, the practice of crafting effective input instructions for LLMs, is a key to using these systems \cite{wei2022chain, brown2020language, reynolds2021prompt}. By careful framing of prompts, users of LLMs can significantly improve the relevance, accuracy, and creativity of generated outputs.  By engineering prompts from data, we can transform abstract metrics into accessible explanations. 

\begin{figure}[t]
    \centering
    \includegraphics[width=\textwidth]{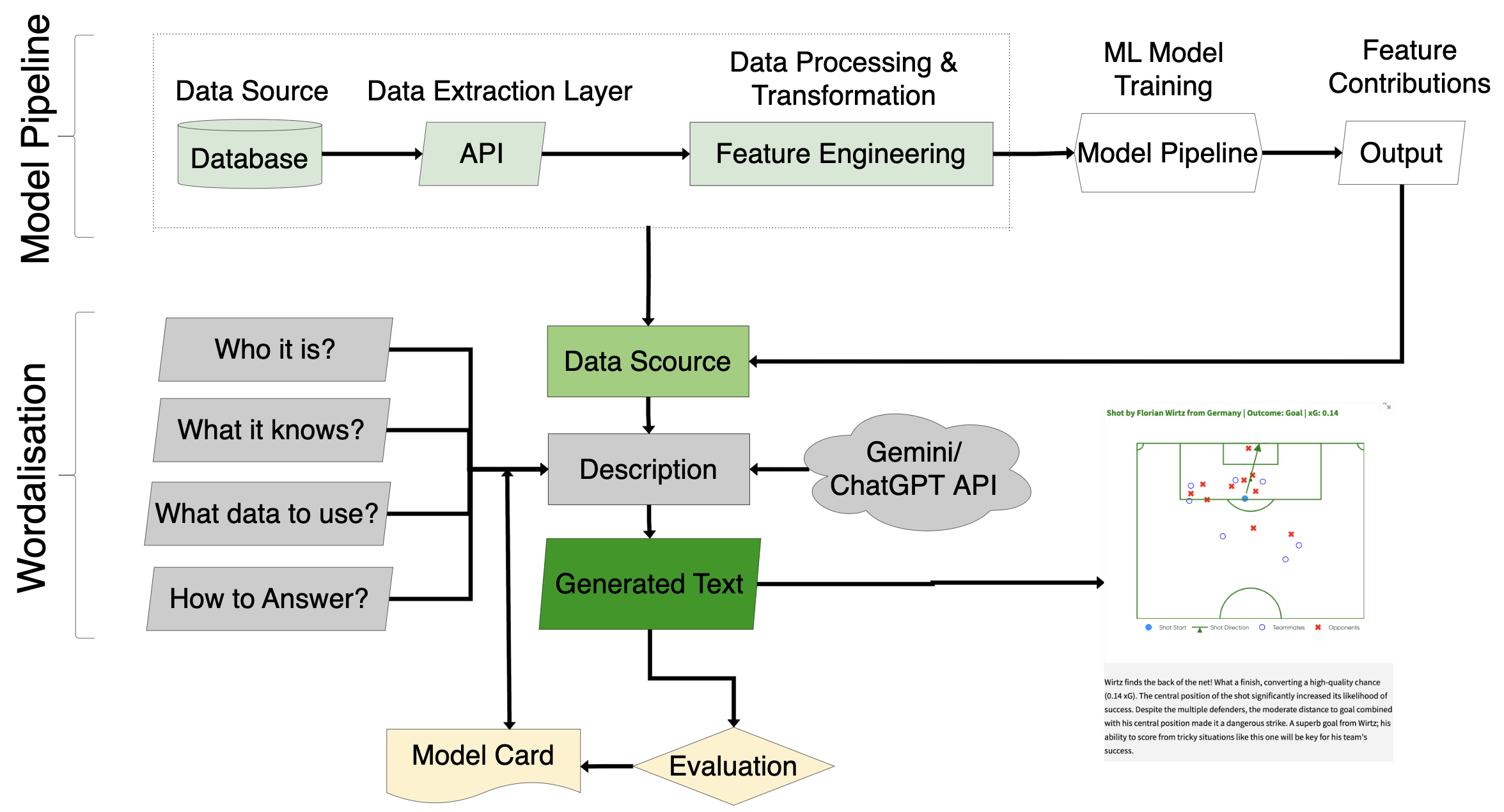}
    \caption{Overview of the proposed workflow, comprising the data pipeline for feature extraction, model training, and feature contribution analysis, along with the wordalisation process that integrates data source, description, and LLM chat modules. The output is an engaging and accurate LLM generated text. }
    \label{fig:wf}
\end{figure}

In this paper, we extend the wordalisation approach to produce texts that interpret models, with expected goals as an example. Figure \ref{fig:wf} illustrates the overall workflow of our proposed approach, which is divided into two main components: the model pipeline and the wordalisation process. The data pipeline extracts data from databases and APIs, generates relevant features, trains corresponding machine learning models, and calculates the contribution of each feature to the output. The second component, wordalisation, uses LLMs to generate intuitive, text-based narratives that explain xG values based on feature contributions. To document our approach, we provide a structured model card \cite{mitchell2019model} detailing design, capabilities, and limitations for transparency and reproducibility. We provide an open-source Streamlit application that enables users to import their own shot dataset and explore xG explanations interactively. The tool is available at \url{https://shotsgpt.streamlit.app/}. We also provide the code online: \url{https://github.com/Peggy4444/shotsGPT/tree/main}.
%\end{itemize}

%By combining explainable AI techniques with LLM-based narratives, this paper advances the interpretability of xG models and paves the way for more accessible and actionable football analytics. We conclude by discussing the system’s limitations, potential applications, and future research directions, including real-world validation with coaching staff.

\section{Methodology}\label{sec3}

We start by describing the dataset and features used to train the xG model. We then explain the model justification and interpretability. Next, we outline the steps for constructing prompts in our wordalisation process. Finally, we introduce metrics to evaluate the wordalisation by analyzing the trade-off between engagement and accuracy.

\subsection{Data Description and Feature Generation}

The dataset used in this study was obtained from the Hudl-StatsBomb events and StatsBomb360 datasets for the following available competitions: EURO Men 2024 and 2022, National Women's Soccer League (NWSL) 2018, FIFA 2022, Women's Super League (FAWSL) 2017, and Africa Cup of Nations (AFCON) 2023 \footnote{\url{https://statsbomb.com/news/statsbomb-release-free-euro-2024-data/}}. These datasets were accessed using the \texttt{statsbombpy} API \footnote{\url{https://github.com/statsbomb/statsbombpy}}. The StatsBomb events dataset comprises 110 columns detailing various aspects of each event, while the StatsBomb360 dataset includes 7 columns describing the positions of players visible in the frame of the action. These datasets were merged to provide a comprehensive view of the events for all matches played by all teams participating in the respective competitions and seasons.

From the shot data, we generated a set of features categorized into body part-related, play pattern-related, goalkeeper-related, shot location-related, and opponent-related features. These features collectively provide a detailed understanding of the shot and its surrounding context. We assume a fixed pitch size of 105 meters in length and 68 meters in width for all games. 

The body part-related features include a binary indicator for whether the shot was taken with the left foot, called \texttt{shot with left foot}. Play pattern-related features consist of binary indicators for \texttt{shots after throw-ins}, \texttt{corners}, and \texttt{free kicks}. Goalkeeper-related features include the Euclidean distance between the shot location and the goalkeeper’s position (\texttt{distance to goalkeeper}), the goalkeeper’s distance to the center of the goal (105, 34) (\texttt{goalkeeper distance to goal}), and the angle between the shot location and the goalkeeper’s position (\texttt{angle to goalkeeper}), measured relative to the goal line. 

\begin{figure}[t!]
    \centering
    \includegraphics[width=\textwidth]{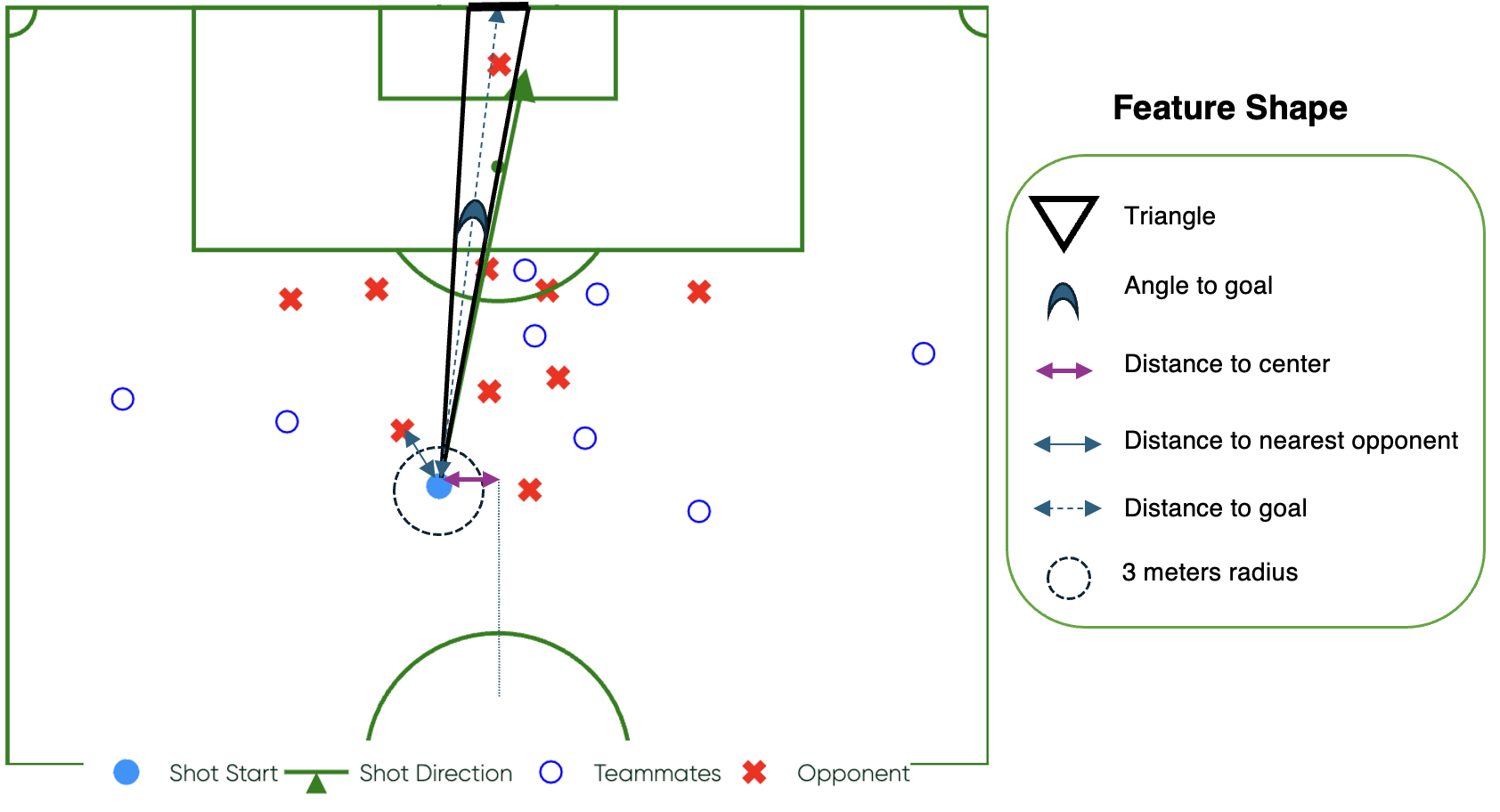}
    \caption{Illustration of various football features including shot location, goalkeeper position, opponent pressure, and teammates' positions.}
    \label{features}
\end{figure}

Shot location-related features encompass the %adjusted x-coordinate of the shot (transformed relative to the goal location), the 
vertical distance from the shot to the centerline of the pitch (\texttt{vertical distance to center}), the angle between the shot location and the goalposts (\texttt{angle to goal}), and the Euclidean distance from the shot location to the goal line (\texttt{distance to goal}). Opponent-related features include the count of opposition players within 3 meters of the ball at the time of the shot (\texttt{nearby opponents in 3 meters}), the number of opponents within a triangular area formed by the shot location and the goalposts (\texttt{opponents in triangle}), the minimum Euclidean distance to the nearest opponent (\texttt{distance to nearest opponent}), and the angle to the nearest opponent from the shot location (\texttt{angle to nearest opponent}). 
These features, which are illustrated in Figure \ref{features}, covers shot positions and position of opposing team, all explaining the success factors of shots.

\subsection{Feature Selection and Model Training}

Our data pipeline is designed to fit a model to that data and build wordalisations which compare the importance of features of a particular shot compared to other shots in that dataset. This means we fit different expected goal models independently to each of the six competitions (to see the individual models go in to \url{https://shotsgpt.streamlit.app/shots} and select the competition of interest from drop-down in the left sidebar and look at the summary table of the trained model for the selected competition). Our aim is not to build the ``best" expected goals model but to be able to explain the probability of scoring a particular shot relative to other shots in the same competition. Depending upon the competition we expect different variables to have different weights in the final model. 

In designing this pipeline, we used one of the competitions (Euros 2024) to identify if any of the variables could be dropped. We found that \texttt{angle to goalkeeper} and \texttt{angle to goal} were highly correlated (Pearson correlation R=0.88), so we dropped the second of these. Similarly, \texttt{distance to goalkeeper} was also highly correlated with \texttt{distance to goal} (Pearson correlation R=0.81), so we dropped the first of these. We then fitted the logistic regression model and looked at the P-values for each of the remaining variables. Among those, \texttt{goalkeeper distance to goal} and \texttt{angle to the nearest opponent} had P-values higher than 0.05, so we dropped them from the model. However, we found that squaring the \texttt{vertical distance to center} made it a significantly explanatory variable, so included it in the model, calling it \texttt{squared distance to center}.

While accuracy is a concern when building predictive models, for our study interpretability is emphasised because we want to build wordalisations to explain the underlying factors that contribute to a shot's outcome. For this reason, we maintain certain features in the model even if their p-value is greater than 0.05, as they provide valuable insights into the shot context. For instance, features like \texttt{shot after throw in} or \texttt{nearby opponents} are kept in the model, as they help explain the circumstances around the shot. 

After this feature selection and transformation process, we arrive at the final set of features that are either (or both) statistically significant and interpretable. The features retained for the final xG model are as follows: \texttt{squared distance to center, euclidean distance to goal, nearby opponents in 3 meters, opponents in triangle, goalkeeper distance to goal, distance to nearest opponent, angle to goalkeeper, shot with left foot, shot after throw in, shot after corner, shot after free-kick}. The same features are included for every competition, although the coefficients vary since they are estimated per competition.

\subsection{Explainable Components and Feature Contribution Weights}

In logistic regression, the predicted probability of an event is modeled as a function of the input features using log-odds. The log-odds can be expressed as a linear combination of the input features, where each feature contributes to the final prediction based on its coefficient. The log-odds for a given shot (feature vector) are defined as:

\begin{equation}
\text{log-odds}(\mathbf{x}) = \beta_0 + \sum_{j=1}^{M} \beta_j x_j, \label{eq:logodds}
\end{equation}
where \( \beta_0 \) is the intercept (baseline),
 \( \beta_j \) is the coefficient for feature \( x_{ij} \),
 and \( x_j \) is the value of feature \( j \). 

We calculate the contribution of each feature to the log-odds by first mean-centering the feature values. This step adjusts each feature value \( x_j \) by subtracting the mean of that feature across the dataset, ensuring that each feature’s contribution is measured relative to its baseline value. The mean-centered feature value for a given shot is denoted as \( \tilde{x}_j \), calculated as:

\[
\tilde{x}_j = x_j - \mu_j,
\]
where \( \mu_j \) is the mean of feature \( x_j \) across all feature vectors (ie. shots) in the dataset.

\begin{figure}[t!]
    \centering
    
    % First row - 56th minute shot
    \begin{subfigure}[b]{0.45\textwidth}
        \centering
        \includegraphics[width=\textwidth]{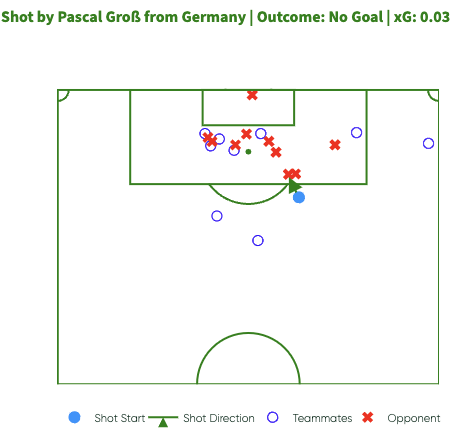}
        \caption{56th minute shot - Pitch Visual}
        \label{fig:field_51}
    \end{subfigure}
    \hfill
    \begin{subfigure}[b]{0.45\textwidth}
        \centering
        \includegraphics[width=\textwidth]{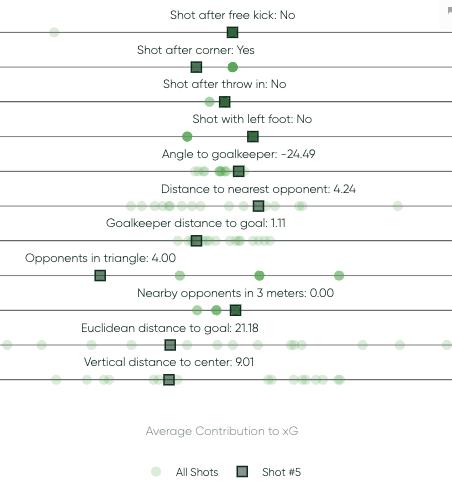}
        \caption{56th minute shot - Contribution Plot}
        \label{fig:contribution_51}
    \end{subfigure}
    
    \vspace{1cm} % Space between rows
    
    % Second row - 85th minute shot
    \begin{subfigure}[b]{0.45\textwidth}
        \centering
        \includegraphics[width=\textwidth]{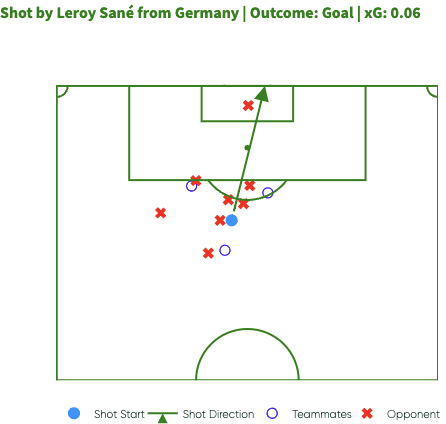}
        \caption{85th minute shot - Pitch Visual}
        \label{fig:field_56}
    \end{subfigure}
    \hfill
    \begin{subfigure}[b]{0.45\textwidth}
        \centering
        \includegraphics[width=\textwidth]{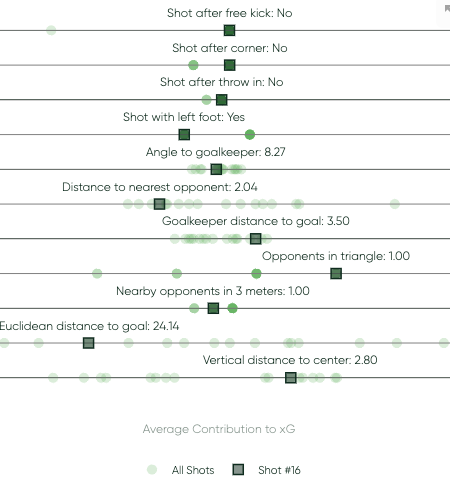}
        \caption{85th minute shot - Contribution Plot}
        \label{fig:contribution_56}
    \end{subfigure}
    
    \caption{Analysis of two shots from Germany vs. Scotland in EURO 2024. The top row shows the 56th-minute shot, with the pitch visual on the left and the contribution plot on the right. The bottom row shows the 85th-minute shot, with the pitch visual on the left and the contribution plot on the right.}
    \label{fig:analysis_shots}
\end{figure}

The contribution of each feature \( x_j \) to the shot's predicted xG is then calculated as

\begin{equation}
\text{Contribution of } x_j = \beta_j \cdot \tilde{x}_j.\label{eq:contrib}
\end{equation}
Here, \( \beta_j \) is the coefficient associated with feature \( x_j \), and \( \tilde{x}_j \) is the mean-centered value of that feature for the specific shot. This approach isolates the unique effect of each feature relative to other observations in the data set. In the context of expected goals, the contribution tells us what was unusual (or not) about this particular shot relative to the other shots in the dataset.

Note that the log-odds are converted to a probability via the logistic function, yielding the final predicted xG:

\begin{equation}
P(y=1 \mid \mathbf{x}) = \frac{1}{1 + e^{-\text{log-odds}(\mathbf{x})}}.
\label{eq:xGprob}
\end{equation}
giving the overall probability of a shot being a goal, i.e the expected goals value. By calculating the contributions of each feature for every shot, we can understand the specific factors driving the model’s prediction. For example, an (unsuccessful) shot in the 56th minute  from a match between Germany and Scotland in EURO 2024 is shown in figures \ref{fig:field_51} and \ref{fig:contribution_51}), with an xG of 0.03. In this case, there are 4 opponents in the triangle blocking the shot path, leading to a large negative contribution from the \texttt{opponents in triangle} feature. This can be seen in the  distribution plot, where  each point is a single shot contribution (i.e. $\beta_j \cdot \tilde{x}_j$) for each of the model variables. 
The fact that the shooter is closely marked by opponents, which significantly reduces the chances of scoring, is thus reflected in the plot, where the value for \texttt{opponents in triangle} is far to the left, indicating a strong negative influence on the xG.

In contrast, the second successful shot, in Figures \ref{fig:field_56} and \ref{fig:contribution_56}, features a slightly higher xG than the first shot. In this instance, there is only one opponent (the goalkeeper) in the triangle blocking the path, and this results in a positive contribution to the xG. The distribution plot shows the \texttt{opponents in triangle} feature far to the right, indicating a positive impact on the xG.

\subsection{Wordalisation: Step by Step Prompt}

\label{stepbystep}

While the approach above explains shot success in terms of the variables, such as defensive pressure (measured by the number of opponents in the shooting triangle), these do not automatically allow communication with practitioners. Visualizations like distribution plots can fall short in conveying actionable or intuitive understanding to coaches, players, or non-technical stakeholders. To address this gap, we adapt the wordalisation approach of \cite{wordalisation}, described in the introduction. 

\begin{figure}[t]
    \centering
    \includegraphics[width=\textwidth]{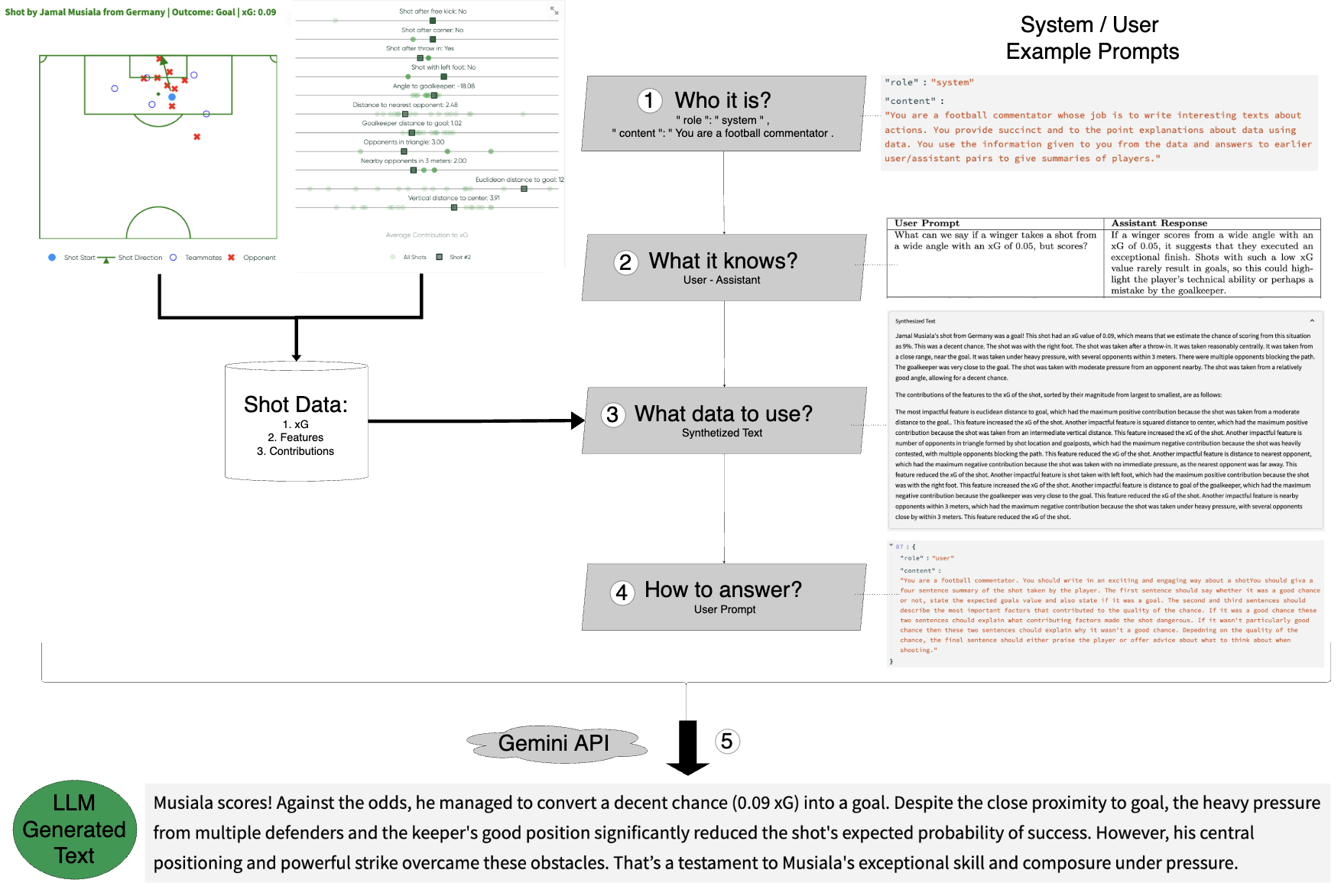}
    \caption{Wordalisation workflow for shots}
    \label{fig:wordalisation}
\end{figure}

There is a structured, four-step approach for creating prompts underlying wordalisations. Each step is designed to provide clarity and context, ensuring the generated descriptions are coherent, aligned with practitioner needs and accurate. These steps are as follows: 1) Tell it who it is, 2) Tell it what it knows, 3) Tell it what data to use, 4) Tell it how to answer. In our case, the aim is to describe these steps tailored for interpreting the contributions of different variables to estimated xG values. An overview of the approach is given in figure \ref{fig:wordalisation}. We now outline the four Wordalisation steps, for, what we call, a shot commentator.

%\begin{enumerate}
 %   \item \textbf{Tell it who it is:} Define the system's role through the system prompt. This step specifies the agent’s identity and purpose, such as acting as a data analyst, scout or commentator.

 %   \item \textbf{Tell it what it knows:} Provide background knowledge using user-assistant question-answer pairs. These instructions shape the model's behavior, offering guidance on how to approach responses and supplying relevant context.

 %   \item \textbf{Tell it what data to use:} Convert the dataset and the selected data point into natural language. This involves translating structured data into words, making the input accessible to the LLM.

 %   \item \textbf{Tell it how to answer:} Offer examples of the desired narrative style. These examples guide the model in crafting clear and actionable descriptions, ensuring the Wordalisation outputs are consistent with user expectations.

%\end{enumerate}

{\bf Tell it who it is: } a large language model's \textbf{system prompt} establishes the context by specifying the role the assistant should fulfill when generating responses. In the case of our shot commentator, we use the system prompt shown in the top right box of \ref{fig:wordalisation}. 

{\bf Tell it what it knows:} The next step involves defining the assistant's knowledge base through example question-and-answer pairs. These examples help the language model understand both the domain-specific knowledge it should convey and the style in which it should respond. An example is shown in figure \ref{fig:wordalisation}. In total we provide 43 question/answer pairs\footnote{https://github.com/soccermatics/twelve-gpt-educational/blob/shots/data/describe/action/shots.xlsx}.

{\bf Tell it what data to use}: The next step is to convert the numerical values of the overall expected goal value (i.e. equation \eqref{eq:xGprob}) and the individual contributions (i.e. equation \eqref{eq:contrib}) into words. This is a very sensitive stage in creating the wordalisation, since it requires us to explain to a coach without a mathematical background, what these equations tell us about football. At this stage, it does not, for example, suffice to simply print out the variable values or the contributions. We need to carefully explain what those values imply about football. 

The xG value (from equation \eqref{eq:xGprob}) quantifies the likelihood of scoring. Instead of giving a numerical value, we use percentiles. Specifically, we translate xG values into qualitative descriptions of scoring chances. We categorize the xG values based on predefined percentiles into five categories: "slim chance" for the 25th percentile (\(< 0.028\)xG ), "low chance" for the 50th percentile (\(< 0.056\)xG), "decent chance" for the 75th percentile (\(< 0.096\)xG), "high-quality chance" for the 90th percentile (\(< 0.3\)xG), and "excellent chance" for values above the 90th percentile (\(> 0.3\)xG). An example of the outputted text is shown in blue in figure \ref{fig:synthesized-text1}.

We also use percentile ranges to describe continuous features such as \texttt{euclidean distance to goal} and \texttt{angle to goalkeeper}, grouping them into categories like "close-range" or "tight angle." For binary features, we employ  mappings, such as: 
\begin{quote}
    "The shot was taken with the left foot." if the value of \texttt{shot with left foot} feature is True.
\end{quote}
An example of these contributions is shown in red in figure~\ref{fig:synthesized-text1}.

\begin{figure}[t!]
    \centering
    
    % Left subfigure
    \begin{subfigure}[t]{\textwidth}
        \centering
        \fbox{
            \begin{minipage}{\textwidth}
                \raggedright
                \textcolor{blue}{\texttt{"Florian Wirtz's shot from Germany was a goal! This shot had an xG value of 0.14, which means that we estimate the chance of scoring from this situation as 14\%. This was a high-quality chance, with a good probability of scoring."}} \\
                \textcolor{red}{\texttt{"The shot was with the right foot. The shot was taken after a throw-in. It was taken from very close to the center of the pitch. It was taken from a moderate distance from the goal. It was taken with moderate pressure, with one opponent within 3 meters. There were multiple opponents blocking the path. The goalkeeper was very close to the goal. The shot was taken with no immediate pressure from any close opponent, with the nearest opponent far away. The shot was taken from a relatively good angle, allowing for a decent chance."}} \\
                \textcolor{darkgray}{\texttt{"The contributions of the features to the xG of the shot, sorted by their magnitude from largest to smallest, are as follows: The most impactful feature is vertical distance to center, which had the maximum positive contribution because the shot was taken closer to the center of the pitch (less vertical distance). This feature increased the xG of the shot. Another impactful feature is euclidean distance to goal, which had the maximum positive contribution because the shot was taken from a moderate distance to the goal. This feature increased the xG of the shot. Another impactful feature is number of opponents in triangle formed by shot location and goalposts, which had the maximum positive contribution because the shot was heavily contested, with multiple opponents blocking the path. This feature increased the xG of the shot. Another impactful feature is shot taken with left foot, which had the maximum positive contribution because the shot was with the right foot. This feature increased the xG of the shot."}}
            \end{minipage}
        }
        \caption{Synthesized text. The text highlighted in blue is the initial description of the expected goals. The red text explains the features in footballing terms, and the grey text describes the feature contributions.}
        \label{fig:synthesized-text1}
    \end{subfigure}
    \hfill
    % Right subfigure
    \begin{subfigure}[t]{\textwidth}
        \centering
        \fbox{
            \begin{minipage}{\textwidth}
                \raggedright
                \texttt{What a strike from Wirtz! Turning a high quality (0.15 xG) chance into a stunning goal. He opened up a position for himself in a central position on the pitch, where he had the time and space to carefully place his shot. Despite multiple opponents between him and the goal, he executed expertly. Wirtz's clinical finishing turned a throw-in into a valuable goal for his team.}
            \end{minipage}
        }
        \caption{Few-shot example.}
        \label{fig:example}
    \end{subfigure}
    
    \caption{Example synthesized text and few-shot example prompt.}
    \label{fig:combined_figures}
\end{figure}

To then explain how different factors contributed to the xG value, we use the contributions as shown in figure \ref{fig:analysis_shots}. By ranking these contributions, positive factors (e.g., "close proximity to the goal") and negative factors (e.g., "poor shooting angle") are highlighted in the text.
For example, a shot with a contribution :
\begin{quote}
    "The high chance of scoring was influenced by the player's close position to the goal and minimal defensive pressure."
\end{quote}
Only features with contributions greater than 0.1 or less than -0.1  in the log-odds are included in these descriptions, as they are more likely to influence the outcome of the shot prediction. This is a somewhat arbitrary choice of threshold, but is motivated by the fact that contributions within the range of $-0.1$ to $0.1$ typically result in only small shifts in the log-odds and, consequently, the xG probability. A full list of such functions used for assigning contributions can be found in the description class of our code \url{https://github.com/Peggy4444/shotsGPT/blob/main/classes/description.py}. The grey text in figure \ref{fig:synthesized-text1} explains the impact of individual features on the xG value, ranked by their contribution magnitude. 

%To further clarify the structure, the synthesized text in Figure~\ref{fig:synthesized-text1} is color-coded: Blue text represents the Shot Quality Description, Red text represents the Shot Feature Description, and Gray text represents the Feature Contribution Description.

{\bf Tell it how to answer:} This stage is focused on crafting the specific instructions and examples that guide the LLM in generating more engaging responses from synthetized text. It gives very specific instructions about the type of text we would like to generate, specifying how many sentences should be written and what each sentence should contain. In addition, we include explicit human-generated examples, a technique known as few-shot prompting \cite{schulhoff2024prompt}. Figure \ref{fig:example} presents a human-written example of a few-shot prompt for the synthesized text shown in figure \ref{fig:synthesized-text1}. We provide three training examples of this type for the wordalisation.

%\clearpage

\subsection{Engagement and Accuracy}

We do automated evaluation of our wordalisations based on two key criteria: engagement and accuracy. We compare five distinct cases. For case 1, the text provided to the evaluation (denoted as [Case text] in the evaluation prompts below) consists only of \textcolor{blue}{shot quality} and \textcolor{red}{features} only (as shown as coloured texts in figure \ref{fig:synthesized-text1}). The idea is to test whether the LLM (Gemini in the examples used here) already has the ability to assess shot value just from a description of the shot, but without additional data. Case 2 extends the text provided in the evaluation prompt to include \textcolor{darkgray}{contributions} (as well as \textcolor{blue}{shot quality} and \textcolor{red}{features}). This provides a comprehensive explanation of the shot and the factors influencing its quality. Case 2 tests the engagement and accuracy of a purely descriptive text.

Cases 3 and 4 test the wordalisations. Case 4 produces a text following the complete wordalisation approach by following all the steps described in section \ref{stepbystep}. Case 3 omits the `tell it what it knows' and `tell it how to answer' stages, to help assess how important these parts of the prompts are in shaping an accurate answer. Finally, case 5 serves as a baseline, providing only numerical feature values without any textual explanation or narrative. 

The aim of our \textbf{engagement} evaluation is to measure how interesting the generated descriptions are to readers.  To calculate engagement using an LLM, we first provide the text we want to evaluate, then ask \textit{``Rank this text on a scale from 0 to 5 for how interesting and engaging it is.''}
This process is repeated for all shots, and the engagement score for each description is then averaged. To ensure robustness, the system includes error handling mechanisms in case of failed responses, retrying the request multiple times.

For \textbf{accuracy} we evaluate how well the generated descriptions align with the true contribution of individual features to the expected goals (xG) value. To do this, an LLM  is provided with a prompt that asks it to assess whether a particular feature (such as Euclidean distance to goal or vertical distance to center) is a positive, negative, or neutral contributor to the xG value. Specifically, we write the prompt
\textit{``In the following text [Case text] was [Feature] a positive, negative, or not contributing factor? Respond with one of [’positive’, ’negative’, ’not contributing’]''}
The output labels are then compared to the ground truth, where features are considered positive if their contribution exceeds 0.1, negative if it is below -0.1, and neutral (not contributing) if it lies between -0.1 and 0.1. The accuracy score is calculated as the percentage of correct assessments made by the LLM across all shot descriptions.

\section{Results}

\subsection{Shot description application}

In order to demonstrate our approach we built a shot description application in Streamlit \url{https://shotsgpt.streamlit.app/}. The application allows the user to select a match from one of the available tournaments, then a shot from that match and it compares the selected shot to the other shots in the match in a distribution plot, shows the location of players and the ball in that shot and writes a short commentry about the shot. The application also allows the user to see the steps used in building the wordalisation: the model summary of the fitted logistic regression; the synthesised text at the "tell it what data to use" stage; and the full sequence of messages sent to the language model. We provide the full code for this application on Github: \url{https://github.com/Peggy4444/shotsGPT/tree/main}.

\subsection{Feature Contributions}

\begin{figure}[t!]
    \centering
    
    % First subfigure (top)
    \begin{subfigure}[b]{\textwidth}
        \centering
        \includegraphics[width=\textwidth]{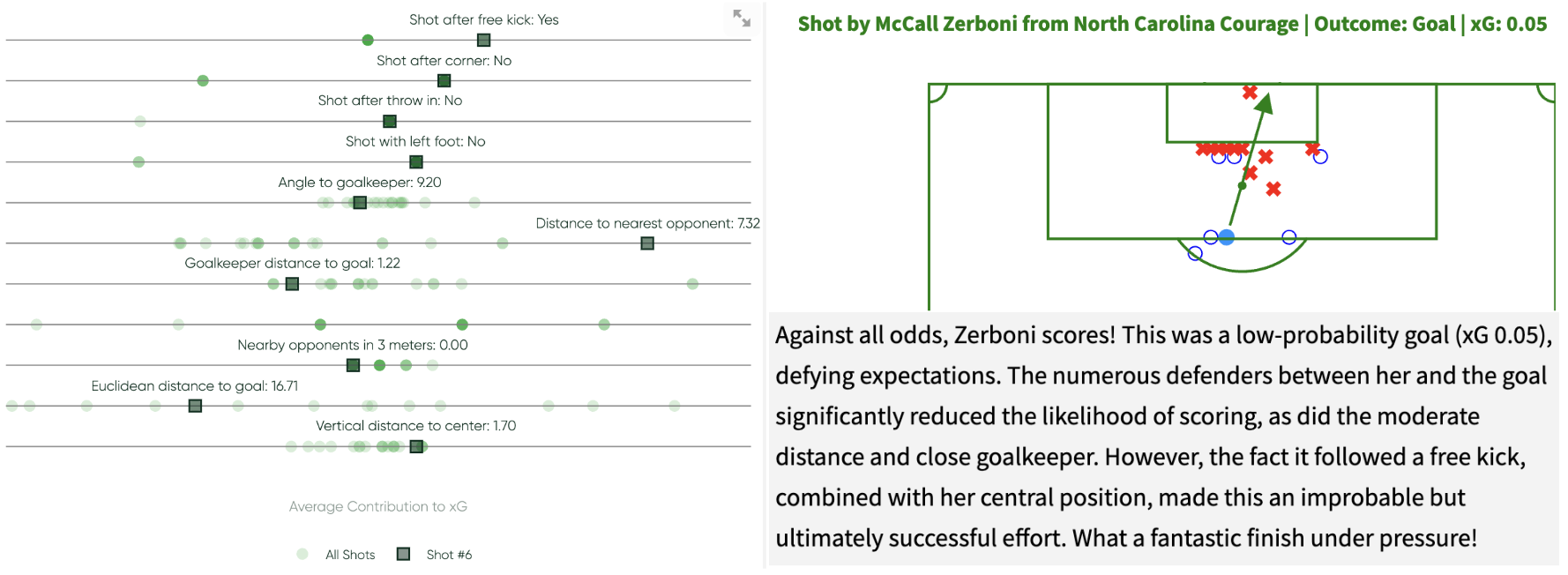}
        \caption{2nd minute shot by McCall Zerboni}
        \label{fig:zerboni}
    \end{subfigure}
    
    % Vertical space between subfigures
    \vspace{1em}
    
    % Second subfigure (bottom)
    \begin{subfigure}[b]{\textwidth}
        \centering
        \includegraphics[width=\textwidth]{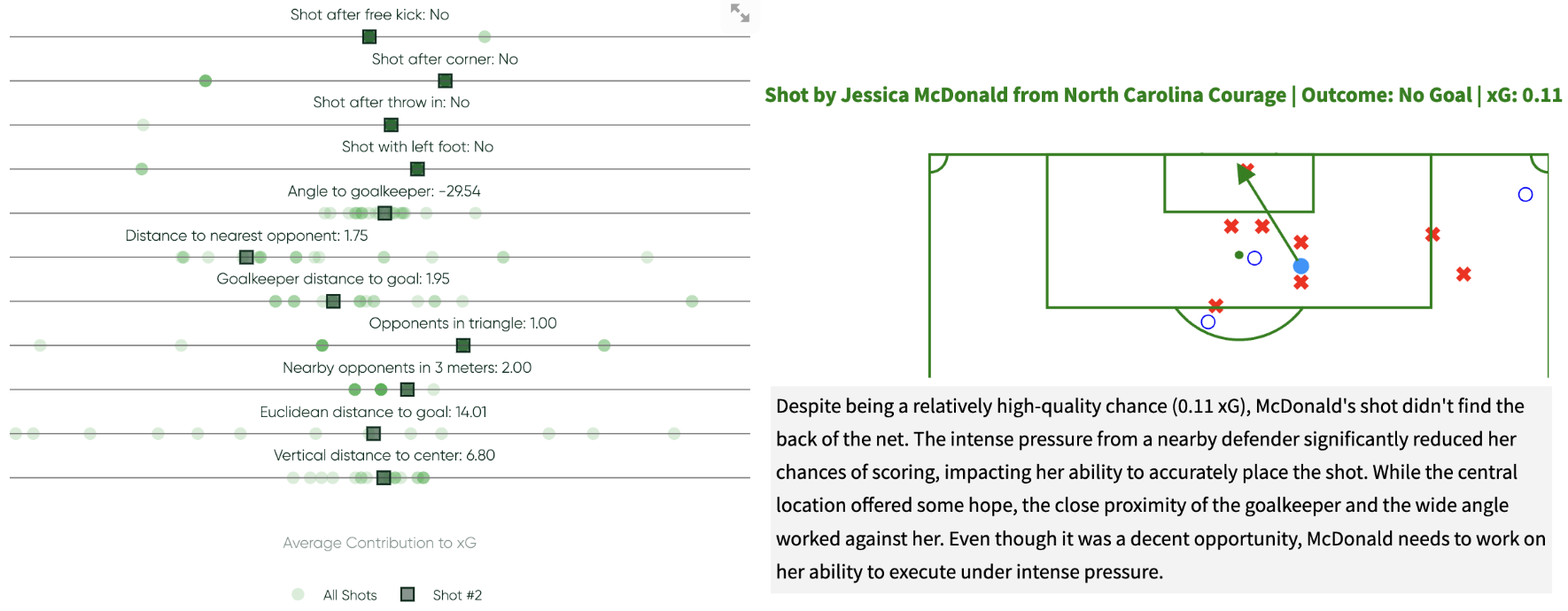}
        \caption{10th minute shot by Jessica McDonald}
        \label{fig:jessica}
    \end{subfigure}
    
    \caption{Feature contribution analysis and LLM generated text of two shots from Washington Spirit vs. North Carolina Courage in National Women's Soccer League (NWSL) 2018.}
    \label{fig:analysis_shotss}
\end{figure}

The contribution plots visualize feature importance (figure \ref{fig:analysis_shotss}).  Each horizontal band in the plot represents a feature, with its width indicating the magnitude of its contribution. Shots with values to the right of the vertical axis have more xG, while those to the left have less xG. In these plots, \texttt{euclidean distance to goal} and \texttt{vertical distance to center} generally emerge as dominant factors. For these variables, the shots are spread out further on the scale: shots from large distances (far on the left), for example, have a reduced probability of goal, while those at short distance (far to the right) have an increased probability.   Figure \ref{fig:analysis_shotss} highlights two contrasting cases. In figure \ref{fig:zerboni}, the shot (which did result in a goal) had a relatively large \texttt{distance to nearest opponent}, increasing the chance of scoring. Conversely, figure \ref{fig:jessica} illustrates an unsuccessful shot where a small \texttt{distance to nearest opponent} significantly reduced the xG. Notably, in both instances, this opponent proximity feature outweighs the typical dominant variables, underscoring the context-dependent nature of feature contributions. The accompanying LLM-generated analysis aligns with these observations, accurately capturing how situational factors alter the relative impact of features on xG predictions.

\subsection{Model Card}

The model card provides a comprehensive overview of its design, capabilities, and limitations. It details the integration of the xG prediction model and the language model. It also outlines the architecture, training data, evaluation metrics, and ethical considerations, emphasizing transparency and interpretability. It also includes structured prompts for the language model and limitations such as dataset bias and feature sensitivity. For further details, including implementation and interactive exploration, refer to the \url{https://github.com/Peggy4444/shotsGPT/blob/main/model\%20cards/model-card-shot-xG-analysis.md}.

\subsection{Evaluation}

\begin{figure}[t]
    \centering
    \includegraphics[width=\textwidth]{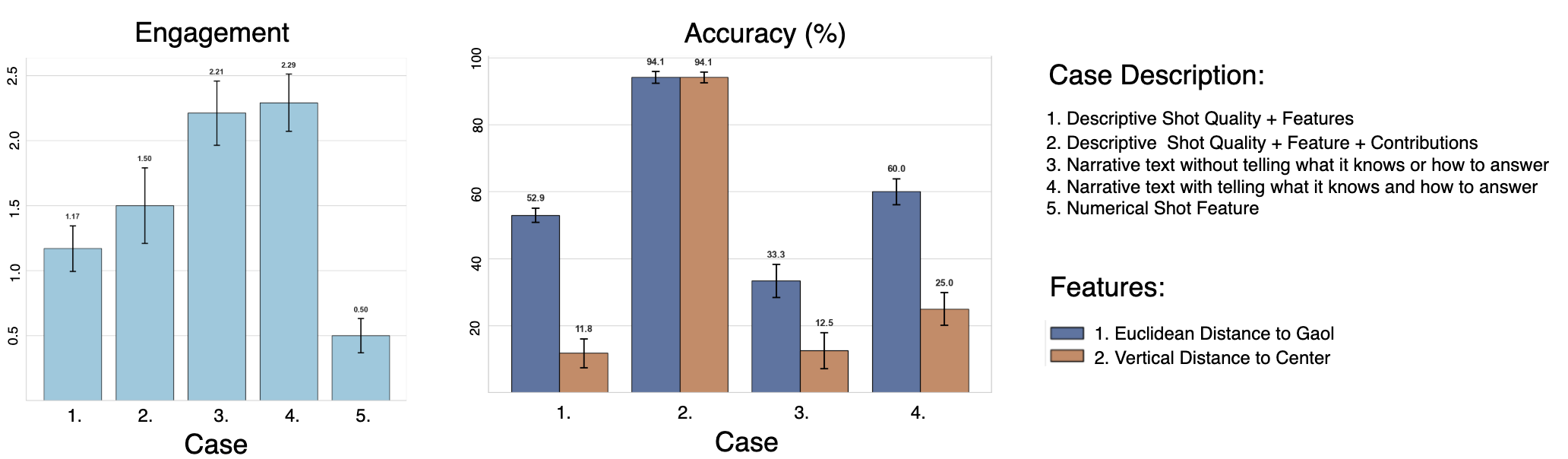}
    \caption{Engagement and Accuracy scores. The results are averaged over 10 runs and standard deviation is shown on top of the bars.}
    \label{fig:eval}
\end{figure}
 Figure~\ref{fig:eval} shows that there is a trade-off between engagement and accuracy in the generated descriptions. Case 2 (i.e., descriptive shot quality + features + contributions) achieves the highest accuracy, for key features such as \texttt{Euclidean distance to goal} and \texttt{vertical distance to center}, as it explicitly lists and explains feature contributions. This  makes it easier for the LLM to identify correct contribution labels. These two features were chosen for accuracy evaluation because they exhibit the strongest influence on xG values, as evidenced by their wider distribution in the contribution plot (Figure~\ref{fig:analysis_shots}). However, Case 2’s engagement score is low. In contrast, case 4 (i.e., full wordalisation), which leverages contextual examples and narrative elements, strikes an optimal balance between accuracy and engagement. It achieves the second-highest accuracy while maintaining the highest engagement score, making it the most suitable for practical use. This balance ensures that the generated descriptions are not only reliable and explainable but also accessible and engaging for football practitioners. The results demonstrate that Case 4 effectively addresses the needs of analysts and coaches by providing insightful, interpretable, and actionable insights into xG values and feature contributions.

\subsection{Further applications}\label{sec:applications}

\begin{figure}[t]
    \centering
    \includegraphics[width=0.9\textwidth]{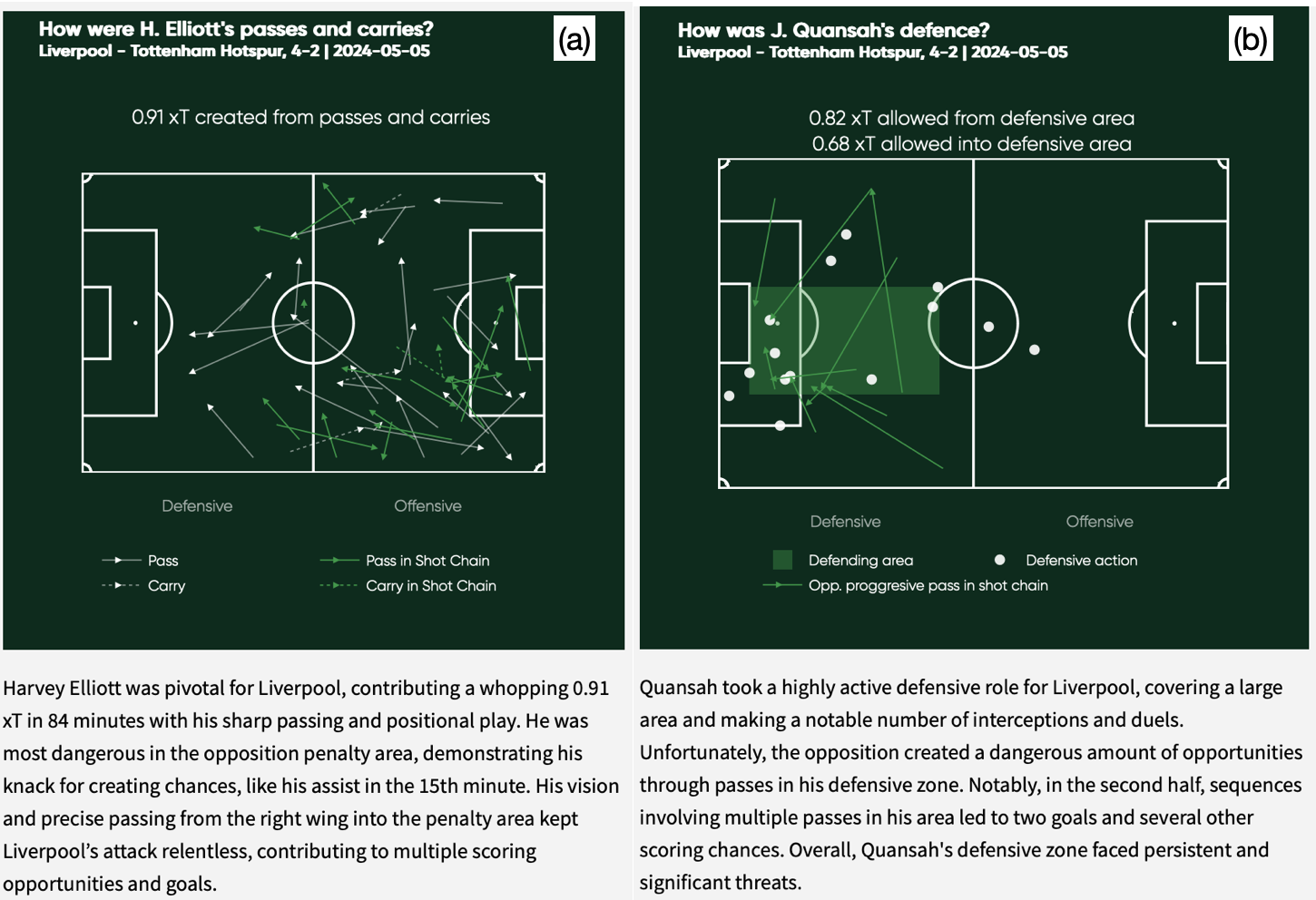}        
    \caption{Wordalisation applications in passes, carries and defensive actions.
    \label{fig:liverpool}}
\end{figure}

So far we have applied this method to evaluating shots, but the same concept can be used to evaluate other actions in football. An example is shown in Figure \ref{fig:liverpool} for both attacking (\ref{fig:liverpool}a) and defensive (\ref{fig:liverpool}b) actions. For attack, we assign an ``exepected threat" value to every pass and carry made by a player and use this to describe the most common type of pass the player makes.  The first step in this process was to create an action-based expected threat model \cite{soccermatics_xtaction}, using three seasons of event data across the French, English, German, Spanish and Italian leagues. This model is a logistic regression predicting the probability that a pass, will eventually be part of a chain of actions leading to a goal.
The model is then interpreted in the context of individual players by creating synthetic descriptions which summarises the  best passes and carries by a player, and also including details of where those passes occurred on the pitch. These synthesized texts are then passed to GPT4o along with both "Tell It What It Knows" question-answer pairs and "Tell It How To Answer" examples. The output is an engaging text, explaining how the player's passes contribute to the team.

\section{Discussion}\label{sec:conclusion}

There are three steps to the process we have outlined for generating natural language narratives describing football shots. The first is to create a mathematical model of the probability of a shot being a goal, in our case a logistic regression. By focusing on variables which are interpretable, we ensure that, at the second step, we can convert the outcome of this model into words. Neither of these steps uses language models and instead we use "old-fashioned" statistical models to fit an expected goal model to data. The linear nature of the logistic regression ensures that we make a correct interpretation of the variables. The third and final step involves combining the "tell it what data to use" text with a series prompts to produce an engaging text about the shots. The resulting text is both engaging and factually correct. 

In terms of explaining what makes a chance good (or poor), we see our approach as an improvement on the SHAP-based feature importance approach \cite{Anzer2021AGS}. The wordalisations not only retain the model's accuracy but also make its outputs more accessible and actionable for end-users. Our automated evaluation methods show that there is a trade-off between an engaging description and an accurate description of all aspects of the shot. This is to be expected, if a coach were to describe the quality of a shot to a player, we would not expect them to give all details in every description.

Our work contributes to the theoretical foundation of wordalisations \cite{wordalisation}, by extending its application to logistic regression models and LLM-generated explanations. Unlike previous implementations that focused on raw numerical rankings, we ask LLMs to interpret the output of machine learning models. Our approach can be extended to other models, such as the expected threat model in figure \ref{fig:liverpool}. Similarly, any model  --- such as pitch control \cite{Spearman2017} and off-ball runs \cite{Fernandez2020}---  which describes positioing and actions of players  can, by following the three steps outlined here, be converted into an informative wordalisation. 

Broadly speaking, models of football can be divided into two approaches: those which give an explicit description of a mechanisms (as discussed in the previous paragraph) and those which use machine learning to make predictions. Although we don't do so here, our approach can potentially be adapted to a more general machine learning setup through the SHAP  values of a model \cite{lundberg2017unified,Anzer2021AGS}. When introducing SHAP, \cite{lundberg2017unified} let \( g \) denote the explanation model approximating the original predictive function \( f \). The explanation for a prediction \( f(\mathbf{x}) \), where \( \mathbf{x} \) is the input, is modeled using a linear equation:
\[
g(\mathbf{z}') = \phi_0 + \sum_{i=1}^{M} \phi_i z'_i,
\]
where:
- \( \mathbf{z}' \) are the simplified input features,
- \( \phi_0 \) is a baseline value, typically the mean prediction of the model,
- \( \phi_i \) represents the contribution of each feature \( z'_i \) to the model’s prediction. For our logistic regression model, equation \ref{eq:logodds} provides a ready-made $g(\mathbf{z}')$ in the form of log-odds, because it is linear in the features. This is not the case for most machine learning models. The challenge then to building wordalisations for general machine learning models is to select first select the simplified input features and then to automate the production of tests around those features.  \cite{Anzer2021AGS} make the first of these steps for an expected goals model based on xGBoost, the further step of wordalising this approach remains an interesting and open research challenge. An important question, however, is whether such an approach is really needed when a more mechanistic apporach (based on logistic regression) is so effective.

Coaches often require insights that are not only accurate, but also easily digestible and actionable \cite{forcher2024soccer,goes2021unlocking}. Our system provides insight by converting complex numerical outputs into intuitive, text-based narratives that highlight key factors influencing xG values, such as shot distance, angle, and defensive pressure. This could allow coaches or players to quickly grasp why a shot has a high or low xG value, enabling more informed decision-making during training and matches. More work is needed, though. A natural next step, is to extend on the automated evaluation we have done here to look at human evaluation by coaching staff. Do the coaches find these descriptions accurate? And, even more imporantly, are they useful in coaching situations? This will be the focus of the next steps of our work.

In summary, we have taken an approach which emphasizes model explainability, not just in a statistical sense, but also in the sense that our models explain the value of a shot in plain language. We believe that machine learning practitioners should endevour to take this approach, which will further help analysts and coaching staff better utilize data without requiring deep expertise in machine learning.

\bibliographystyle{plainnat}  
%\bibliography{references}  

\begin{thebibliography}{34}
\providecommand{\natexlab}[1]{#1}
\providecommand{\url}[1]{\texttt{#1}}
\expandafter\ifx\csname urlstyle\endcsname\relax
  \providecommand{\doi}[1]{doi: #1}\else
  \providecommand{\doi}{doi: \begingroup \urlstyle{rm}\Url}\fi

\bibitem[Anzer and Bauer(2021)]{Anzer2021AGS}
Gabriel Anzer and Pascal Bauer.
\newblock A goal scoring probability model for shots based on synchronized positional and event data in football (soccer).
\newblock \emph{Frontiers in Sports and Active Living}, 3, 2021.
\newblock URL \url{https://api.semanticscholar.org/CorpusID:232387328}.

\bibitem[Bransen and Davis(2021)]{Bransen2021}
L.~Bransen and J.~Davis.
\newblock Women’s football analyzed: interpretable expected goals models for women.
\newblock In \emph{AI for Sports Analytics (AISA) Workshop at IJCAI 2021}, Montreal, Canada, 2021.

\bibitem[Brown et~al.(2020)Brown, Mann, Ryder, Subbiah, Kaplan, Dhariwal, Neelakantan, Shyam, Sastry, Askell, et~al.]{brown2020language}
Tom~B Brown, Benjamin Mann, Nick Ryder, Melanie Subbiah, Jared Kaplan, Prafulla Dhariwal, Arvind Neelakantan, Pranav Shyam, Girish Sastry, Amanda Askell, et~al.
\newblock Language models are few-shot learners.
\newblock \emph{Advances in neural information processing systems}, 33:\penalty0 1877--1901, 2020.

\bibitem[Caut et~al.(2025)Caut, Rouillard, Zenebe, Green, Ágúst Pálmason~Morthens, and Sumpter]{wordalisation}
Amandine~M. Caut, Amy Rouillard, Beimnet Zenebe, Matthias Green, Ágúst Pálmason~Morthens, and David J.~T. Sumpter.
\newblock Representing data in words.
\newblock https://arxiv.org/abs/2503.15509, 2025.

\bibitem[Davis and Robberechts(2020)]{Davis2020}
Jesse. Davis and Peter. Robberechts.
\newblock How data availability affects the ability to learn good xg models.
\newblock In \emph{Proceedings of the 7th International Workshop of Machine Learning and Data Mining for Sports Analytics}, 2020.

\bibitem[Davis et~al.(2024)]{davis2024challenges}
K.~Davis et~al.
\newblock Challenges in sports analytics: Methodological and evaluation considerations.
\newblock \emph{Machine Learning}, 113:\penalty0 6977--7010, 2024.

\bibitem[Decroos et~al.(2019)Decroos, Bransen, Van~Haaren, and Davis]{Tom2019}
Tom Decroos, Lotte Bransen, Jan Van~Haaren, and Jesse Davis.
\newblock Actions speak louder than goals: Valuing player actions in soccer.
\newblock In \emph{In ACM KDD}, 2019.

\bibitem[Dick and Brefeld(2022)]{Brefeld2022}
Uwe Dick and Ulf Brefeld.
\newblock Action rate models for predicting actions in soccer.
\newblock \emph{AStA Advances in Statistical Analysis}, 2022.

\bibitem[Eggels et~al.(2016)Eggels, Van~Elk, and Pechenizkiy]{Eggels2016}
H.~Eggels, R.~Van~Elk, and M.~Pechenizkiy.
\newblock Explaining soccer match outcomes with goal scoring opportunities predictive analytics.
\newblock In \emph{Proceedings of the Workshop on Machine Learning and Data Mining for Sports Analytics 2016 co-located with the 2016 European Conference on Machine Learning and Principles and Practice of Knowledge Discovery in Databases}, Garda, Italy, 2016.

\bibitem[Fernandez et~al.(2019)Fernandez, Bornn, and Cervone]{Fernandez2019}
Javier Fernandez, Luke Bornn, and Dan Cervone.
\newblock Decomposing the immeasurable sport: A deep learning expected possession value framework for soccer.
\newblock In \emph{13th MIT Sloan Sports Analytics Conference}, 2019.

\bibitem[Forcher et~al.(2024)Forcher, Forcher, and Altmann]{forcher2024soccer}
Leon Forcher, Leander Forcher, and Stefan Altmann.
\newblock How soccer coaches can use data to better develop their players and be more successful.
\newblock In \emph{Individualizing Training Procedures with Wearable Technology}, pages 99--123. Springer, 2024.

\bibitem[Goes et~al.(2021)Goes, Meerhoff, Bueno, Rodrigues, Moura, Brink, Elferink-Gemser, Knobbe, Cunha, Torres, et~al.]{goes2021unlocking}
FR~Goes, LA~Meerhoff, MJO Bueno, DM~Rodrigues, FA~Moura, MS~Brink, MT~Elferink-Gemser, AJ~Knobbe, SA~Cunha, RS~Torres, et~al.
\newblock Unlocking the potential of big data to support tactical performance analysis in professional soccer: A systematic review.
\newblock \emph{European Journal of Sport Science}, 21\penalty0 (4):\penalty0 481--496, 2021.

\bibitem[Gyarmati and Stanojevic(2016)]{gyarmati2016qpass}
Laszlo Gyarmati and Rade Stanojevic.
\newblock A merit-based evaluation of soccer passes.
\newblock In \emph{ACM KDD Workshop on Large-Scale Sports Analytics}, 2016.

\bibitem[Herbinet(2018)]{Herbinet2018}
C.~Herbinet.
\newblock Predicting football results using machine learning techniques.
\newblock Meng thesis, Imperial College London, 2018.

\bibitem[Lucey et~al.(2015)Lucey, Bialkowski, Monfort, Carr, and Matthews]{lucey2015quality}
Patrick Lucey, Alina Bialkowski, Mathew Monfort, Peter Carr, and Iain Matthews.
\newblock quality vs quantity: Improved shot prediction in soccer using strategic features from spatiotemporal data.
\newblock 2015.

\bibitem[Lundberg and Lee(2017)]{lundberg2017unified}
Scott~M Lundberg and Su-In Lee.
\newblock A unified approach to interpreting model predictions.
\newblock \emph{NeurIPS}, 30, 2017.

\bibitem[Mitchell et~al.(2019)Mitchell, Maziarka, Kamar, Caruana, Wallach, Dastin, Friedler, Dastin, Kim, Lou, et~al.]{mitchell2019model}
Margaret Mitchell, Petra Maziarka, Ece Kamar, Rich Caruana, Hanna Wallach, Jeffrey Dastin, Sorelle~A Friedler, Jeffrey Dastin, Ben Kim, Kelsey Lou, et~al.
\newblock Model cards for model reporting.
\newblock \emph{Proceedings of the 2019 CHI Conference on Human Factors in Computing Systems}, pages 1--13, 2019.

\bibitem[Morales(2016)]{morales2016mathematics}
C{\'e}sar~A Morales.
\newblock A mathematics-based new penalty area in football: tackling diving.
\newblock \emph{Journal of sports sciences}, 34\penalty0 (24):\penalty0 2233--2237, 2016.

\bibitem[Pardo(2020)]{Pardo2020}
M.~Pardo.
\newblock Creating a model for expected goals in football using qualitative player information.
\newblock Master's thesis, Universitat Politecnica de Catalunya, 2020.

\bibitem[Peralta~Alguacil et~al.(2020)Peralta~Alguacil, Fernandez, Piñones~Arce, and Sumpter]{Fernandez2020}
Francisco Peralta~Alguacil, Javier Fernandez, Pablo Piñones~Arce, and David Sumpter.
\newblock Seeing in to the future: using self-propelled particle models to aid player decision-making in soccer.
\newblock In \emph{In Proceedings of the 14th MIT Sloan Sports Analytics Conference}, 2020.

\bibitem[Pollard and Reep(1997)]{pollard1997measuring}
Richard Pollard and Charles Reep.
\newblock Measuring the effectiveness of playing strategies at soccer.
\newblock \emph{Journal of the Royal Statistical Society Series D: The Statistician}, 46\penalty0 (4):\penalty0 541--550, 1997.

\bibitem[Rahimian et~al.(2022)Rahimian, Van~Haaren, Abzhanova, and Toka]{pegah2022}
Pegah Rahimian, Jan Van~Haaren, Togzhan Abzhanova, and Laszlo Toka.
\newblock Beyond action valuation: A deep reinforcement learning framework for optimizing player decisions in soccer.
\newblock In \emph{16th MIT Sloan Sports Analytics Conference}, 2022.

\bibitem[Rahimian et~al.(2023)Rahimian, Van~Haaren, and Toka]{pegah2023-2}
Pegah Rahimian, Jan Van~Haaren, and Laszlo Toka.
\newblock Towards maximizing expected possession outcome in soccer.
\newblock \emph{International Journal of Sports Science and Coaching}, 2023.

\bibitem[Rathke(2017)]{Rathke2017AnEO}
Alex A.~T. Rathke.
\newblock An examination of expected goals and shot efficiency in soccer.
\newblock \emph{Journal of Human Sport and Exercise}, 12:\penalty0 514--529, 2017.
\newblock URL \url{https://api.semanticscholar.org/CorpusID:148713007}.

\bibitem[Reynolds and McDonell(2021)]{reynolds2021prompt}
Lariah Reynolds and Kyle McDonell.
\newblock Prompt programming for large language models: Beyond the few-shot paradigm.
\newblock \emph{arXiv preprint arXiv:2102.07350}, 2021.

\bibitem[Ruiz et~al.(2015)Ruiz, Lisboa, Neilson, and Gregson]{Ruiz2015MeasuringSE}
H{\'e}ctor Ruiz, Paulo J.~G. Lisboa, Paul Neilson, and Warren Gregson.
\newblock Measuring scoring efficiency through goal expectancy estimation.
\newblock In \emph{The European Symposium on Artificial Neural Networks}, 2015.

\bibitem[Sarkar and Kamath(2021)]{Sarkar2021}
S.~Sarkar and S.~Kamath.
\newblock Does luck play a role in the determination of the rank positions in football leagues? a study of europe’s big five.
\newblock \emph{Ann Oper Res}, 2021.
\newblock \doi{10.2202/1559-0410.1014}.

\bibitem[Schulhoff et~al.(2024)Schulhoff, Ilie, Balepur, Kahadze, Liu, Si, Li, Gupta, Han, Schulhoff, et~al.]{schulhoff2024prompt}
Sander Schulhoff, Michael Ilie, Nishant Balepur, Konstantine Kahadze, Amanda Liu, Chenglei Si, Yinheng Li, Aayush Gupta, HyoJung Han, Sevien Schulhoff, et~al.
\newblock The prompt report: A systematic survey of prompting techniques.
\newblock \emph{arXiv preprint arXiv:2406.06608}, 2024.

\bibitem[Spearman et~al.(2017)Spearman, Basye, Dick, Hotovy, and Hudl]{Spearman2017}
William~Robert Spearman, Austin~Thomas Basye, Gregory~J. Dick, Ryan Hotovy, and Paul~Pop Hudl.
\newblock Physics-based modeling of pass probabilities in soccer.
\newblock In \emph{MIT Sloan Sports Analytics Conference}, 2017.

\bibitem[Sumpter(2023)]{soccermatics_xtaction}
D.~J.~T. Sumpter.
\newblock Expected threat - action-based.
\newblock \url{https://soccermatics.readthedocs.io/en/latest/lesson4/xTAction.html}, 2023.

\bibitem[Sumpter(2016)]{sumpter2016soccermatics}
David Sumpter.
\newblock \emph{Soccermatics: mathematical adventures in the beautiful game}.
\newblock Bloomsbury Publishing, 2016.

\bibitem[Tippana(2020)]{Tippana2020}
T.~Tippana.
\newblock How accurately does the expected goals model reflect goalscoring and success in football?, 2020.

\bibitem[Wei et~al.(2022)Wei, Wang, Schuurmans, Bosma, Ichter, Xia, Chi, Le, and Zhou]{wei2022chain}
Jason Wei, Xuezhi Wang, Dale Schuurmans, Maarten Bosma, Brian Ichter, Fei Xia, Ed~H Chi, Quoc Le, and Denny Zhou.
\newblock Chain-of-thought prompting elicits reasoning in large language models.
\newblock \emph{arXiv preprint arXiv:2201.11903}, 2022.

\bibitem[Wheatcroft and Sienkiewicz(2021)]{Wheatcroft2021}
E.~Wheatcroft and E.~Sienkiewicz.
\newblock A probabilistic model for predicting shot success in football.
\newblock \emph{arXiv preprint arXiv:2101.02104}, 2021.

\end{thebibliography}

\end{document}